\documentclass[conference]{IEEEtran}
\IEEEoverridecommandlockouts
\usepackage{cite}
\usepackage{amsmath,amssymb,amsfonts}
\usepackage{algorithmic}
\usepackage{graphicx}
\usepackage{graphics}
\usepackage{textcomp}
\usepackage{pgfplots}
\usepackage{xcolor}
\usepackage{colortbl}
\usepackage{booktabs} % Francisco    
\usepackage{subfig} % Francisco
\usepackage{multirow} % Francisco
\usepackage{hyperref} % Francisco
\usepackage{todonotes} % Francisco

\def\BibTeX{{\rm B\kern-.05em{\sc i\kern-.025em b}\kern-.08em
    T\kern-.1667em\lower.7ex\hbox{E}\kern-.125emX}}
\begin{document}

\title{Evaluating the plausibility of synthetic images for improving automated endoscopic stone recognition 
}
\author{
Ruben Gonzalez-Perez$^{\dagger,1}$, Francisco Lopez-Tiro$^{\dagger,1,2}$,
Ivan Reyes-Amezcua$^{3}$, Eduardo Falcon-Morales$^{1}$ \\ Rosa-Maria Rodríguez-Guéant$^{5}$,
 Jacques Hubert$^{4}$, Michel Daudon$^{6}$,
Gilberto Ochoa-Ruiz*$^{,1}$, Christian Daul*$^{,2}$  \\
$^{1}$Tecnologico de Monterrey, School of Sciences and Engineering, Mexico \\ 
$^{2}$CRAN (UMR 7039, Université de Lorraine and CNRS), Nancy, France \\
$^{3}$CINVESTAV, Computer Sciences Department, Guadalajara, Mexico \\
$^{4}$CHRU Nancy, Service d’urologie de Brabois, Vand{\oe}uvre-l\`es-Nancy, France  
\\$^{5}$NGERE - Nutrition-Génétique et Exposition aux Risques Environnementaux Nancy, France \\
$^{6}$H\^opital Tenon, AP-HP, Paris, France 
\thanks{$^{\dagger}$Equal contribution}
\thanks{*Corresponding authors:} %
\thanks{gilberto.ochoa@tec.mx, christian.daul@univ-lorraine.fr}
}

\maketitle

\begin{abstract}
Currently, the Morpho-Constitutional Analysis (MCA) is the de facto approach for the etiological diagnosis of kidney stone formation, and it is an important step for establishing personalized treatment to avoid relapses. More recently, research has focused on performing such tasks intra-operatively, an approach known as Endoscopic Stone Recognition (ESR). Both methods rely on features observed in the surface and the section of kidney stones to separate the analyzed samples into several sub-groups. However, given the high intra-observer variability and the complex operating conditions found in ESR, there is a lot of interest in using AI for computer-aided diagnosis.  However, current AI models require large datasets to attain a good performance and for generalizing to unseen distributions. This is a major problem as large labeled datasets are very difficult to acquire, and some classes of kidney stones are very rare. Thus, in this paper, we present a method based on diffusion as a way of augmenting pre-existing ex-vivo kidney stone datasets. Our aim is to create plausible diverse kidney stone images that can be used for pre-training models using ex-vivo data. We show that by mixing natural and synthetic images of CCD images, it is possible to train models capable of performing very well on unseen intra-operative data. Our results show that is possible to attain an improvement of 10\% in terms of accuracy compared to a baseline model pre-trained only on ImageNet. Moreover, our results show an improvement of 6\% for surface images and 10\% for section images compared to a model train on CCD images only, which demonstrates the effectiveness of using synthetic images.
\end{abstract}


\section{Introduction}
\label{sec:intro}

\begin{figure*}[ht]
  \begin{center}
    \includegraphics[width=0.78\textwidth]{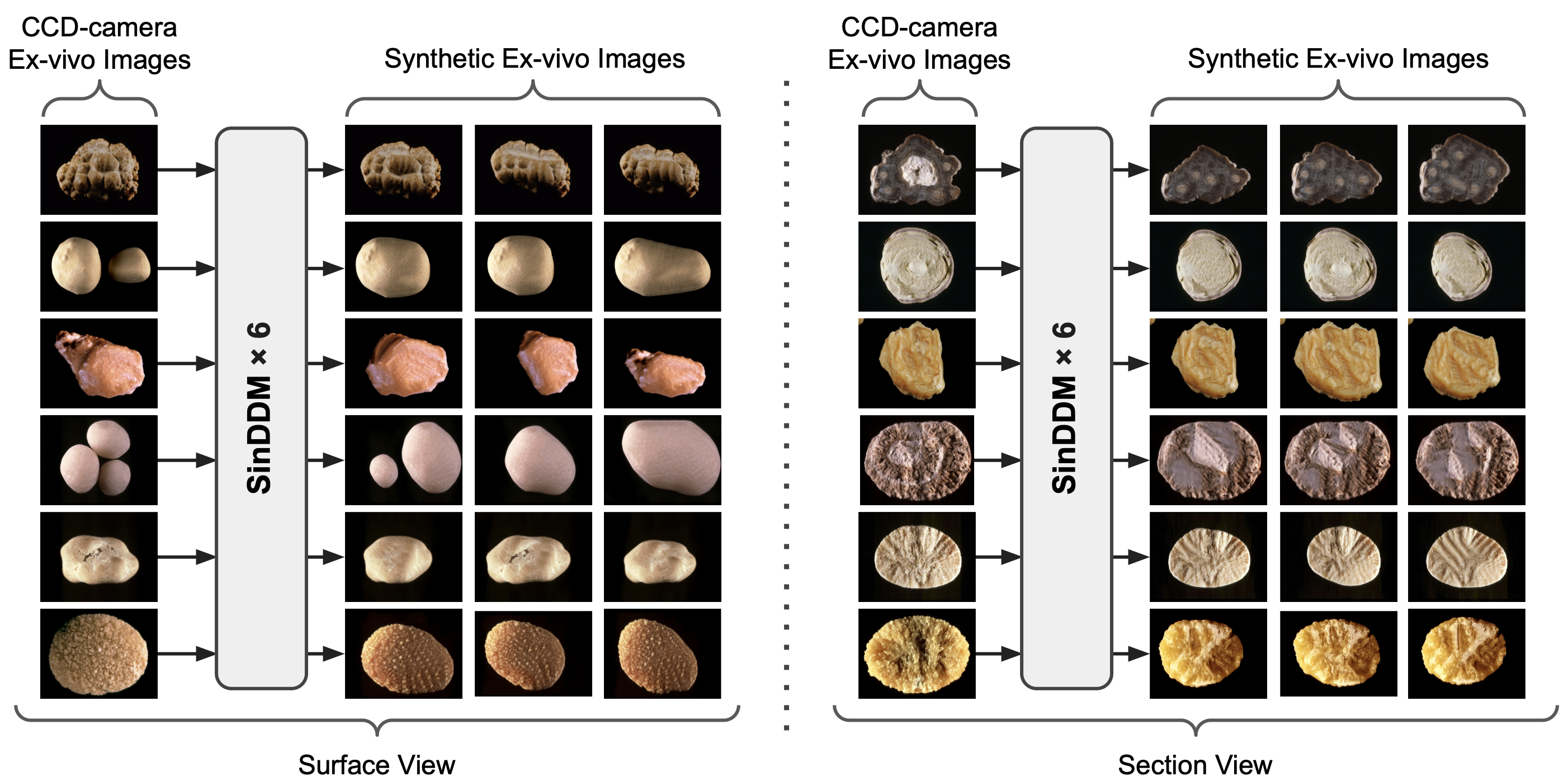}
  \end{center}
\vspace{-0.3cm}
  \caption{Examples of synthetic image generation for surface (external) and section (internal) views of kidney stones. The images were generated using the SinDDM model from a dataset of ex-vivo kidney stone images acquired with a charge-coupled device (CCD) camera. The synthetic images present characteristics highly similar in color and texture to the CCD-camera images.}
  \label{fig:motivation}
\end{figure*}

\subsection{Medical Context}

The use of artificial intelligence (AI) is increasingly gaining importance in medical imaging applications at large. In particular deep learning (DL) methods have been investigated for implementing Computer Aided Diagnosis (CAD) tools \cite{shaikh2021artificial}. %, wang2021review}.
%
%Recently, Deep Learning (DL) methods-based AI techniques have been actively investigated in medical imaging with potential computer-aided diagnosis (CADx) applications  
%
For medical imaging applications in endoscopy and ureteroscopy, there is a great interest in developing methods for the classification, detection, and segmentation in diverse areas of interest, such as early cancer assessment or kidney stone identification, to name a few \cite{ali2022we}. % in which there are some significant advances . 
However, one of the main drawbacks of DL methods is their reliance on large amounts of data for training and testing high-performance models. %\cite{razzak2018deep}. 
This is particularly evident in the medical domain, where it is difficult to gather enough training images of certain diseases or medical conditions \cite{lopez2023boosting}. Another issue is related to domain shift: models trained with data from a given acquisition device will not generalize well when faced with data from another hospital center or equipment vendor. 
%is the lack of enough images to train DL models or its weight initialization  with a similar distribution close to the target domain \cite{lopez2023boosting}, and also synthetic data simulating different acquisition conditions and different types of endoscopes.

This problem is particularly evident in the context of endoscopic examinations for kidney stone identification: the prevalent diagnostic technique for such purpose, known as the Morpho-Constitutional Analysis (MCA) \cite{daudon2016comprehensive}, relies on the visual inspection of extracted kidney stones (surface and section views), in tandem with an FTIR spectral analysis for fine-grained biochemical information. 
%\cite{daudon2012stone}. 
Given the high intra-observer variability of the process, as well as the long processing time in the hospital to obtain the outcome of this study, several researchers have investigated the use of computer vision and deep learning for automating the identification of the extracted samples, either using ex-vivo or in-vivo data \cite{lopez2024vivo}. 

However, most automated MCA systems have been used under strictly controlled acquisition conditions, which might limit their applicability in more realistic scenarios. In addition, the MCA analysis presents large class imbalances, due to the fact that certain types of kidney stones are extremely rare \cite{corrales2021classification}. Such issue makes it difficult to train and validate systems due to model bias, poor performance in the minority class, and inaccurate evaluation of model performance \cite{huynh2022semi}. 

\subsection{Motivation}

 Several recent works have proposed the use of Generative Adversarial Networks (GANs) %\cite{goodfellow2014generative} 
 or more recently, Diffusion Models %\cite{sohldickstein2015deep} 
 to create new samples for copying with data imbalance and out-of-distribution issues in medical systems \cite{vats2023evaluating}. In this contribution, we present the results of a preliminary study for the assessment of synthetically generated images of kidney stones from a pre-existing ex-vivo dataset (Fig. \ref{fig:motivation}). 
 
 This source dataset is comprised of high-resolution charge-coupled device (CCD) images with uniform background and controlled illumination conditions \cite{corrales2021classification}.
The objective of our work is to assess if the use of synthetically generated images can improve the performance of the model in an out-of-distribution setting: our aim is to pre-train deep learning models on ex-vivo data and test the models on in-vivo (i.e., acquired using and ureteroscope) images. The rationale for our proposal is that urologists are increasingly interested in using endoscopic images for performing the visual inspection component of the MCA analysis during the extraction surgery (Endoscopic Stone Recognition or ESR). Such automated procedures for ESR are nowadays vital as urologists are increasingly employing laser lithotripsy for converting the stone into dust, losing valuable visual information for diagnosis in the process \cite{estrade2022towards}. Therefore, our aim in this work is to assess whether the fusion of the source and synthetically generated images for surface and section images improves the overall performance of the model.

\begin{figure*} [] 
    \centering
    \subfloat[Generation of synthetic images]{\label{fig:method_generation}
    \includegraphics[width=0.75\textwidth]{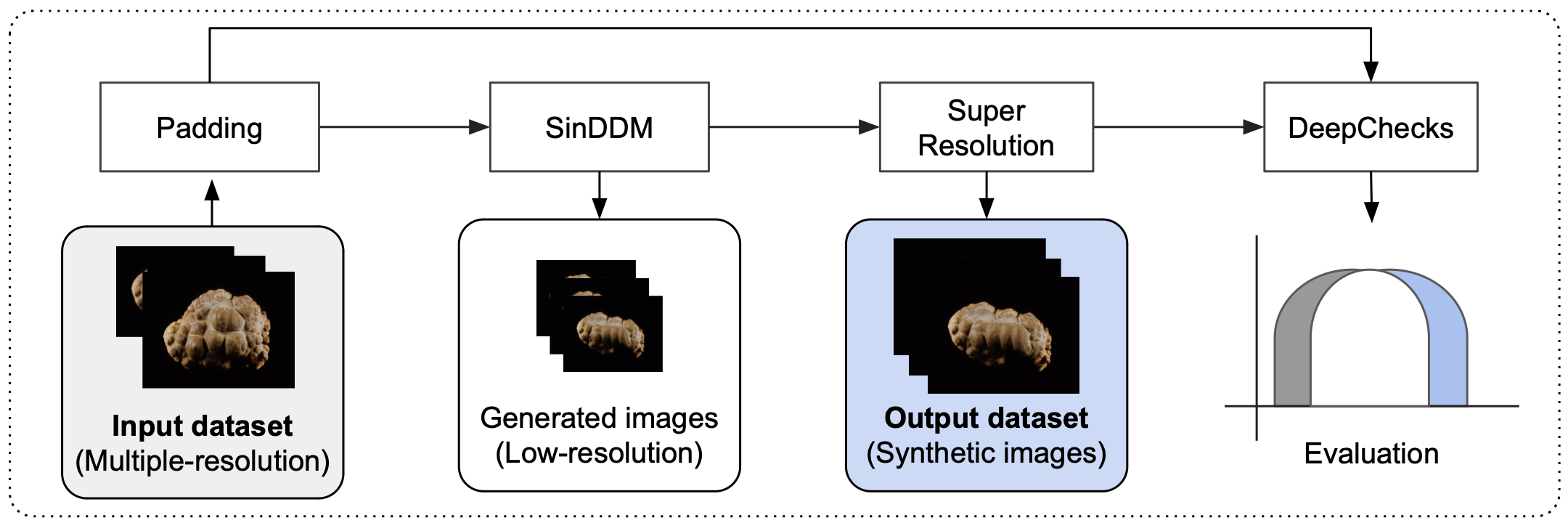}}
\vspace{-0.25cm}
    \subfloat[Classification of kidney stones]{
\label{fig:method_classification}\includegraphics[width=0.75\textwidth]{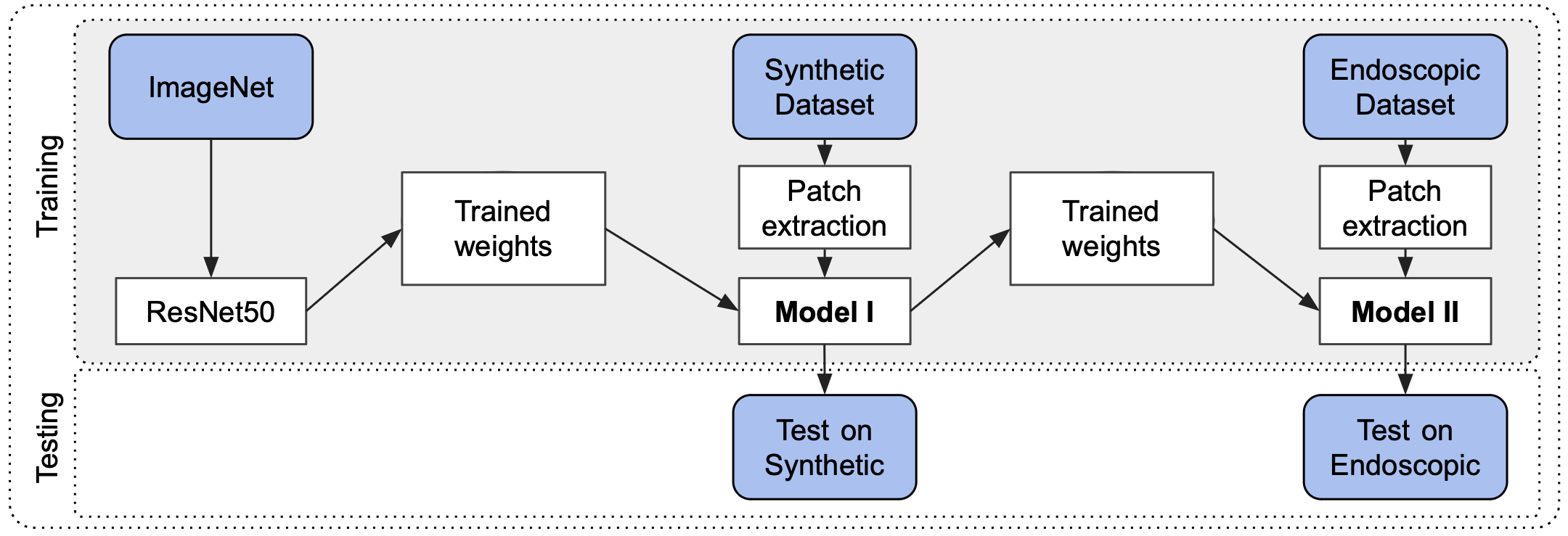}}
\vspace{-0.1cm}
\caption{Evaluating the plausibility of synthetically generated images for automated endoscopic stone recognition pipeline. (a) Generation of synthetic images: The process begins with a set of standard CCD images that are adjusted to dimensions of 2848$\times$4288 pixels through padding. These images are then used to generate low-resolution synthetic versions (264$\times$200 pixels) using SinDDM. Subsequently, these synthetic images undergo an x4 super-resolution process. Finally, the similarity between the input dataset (CCD camera images) and the output dataset (synthetic images) is evaluated using DeepChecks. To carry out the (b) automatic classification of kidney stones in endoscopic images for six different classes, training of models I and II is required. Model I is trained on the synthetic dataset and weights learned from ImageNet (1st TL step). Then, the endoscopic dataset is trained on Model II along with the weights learned in the synthetic distribution (2nd TL step).}  
    \label{fig:method}
    \end{figure*}

\subsection{Contributions of this paper}

Although several methods the been presented in the literature for automating both MCA and ESR, such models have been trained and tested on image patches \cite{lopez2024vivo}. Despite the promising results yielded by these methods, they all present a common problem: insufficient and unbalanced images of samples per class, which makes it difficult to train and validate DL models. Although solutions such as patch-based data augmentation have been proposed \cite{lopez2024vivo}, it is more clinically realistic to test such models on whole images, and not on portions of the same image, which does not match the procedure followed by experts to perform the visual inspection of MCA. In order to train and validate models on complete images, in this contribution, a preliminary approach for the generation of synthetic images of kidney stones in the style of images acquired with a standard CCD-camera is presented. 

To accomplish this, herein we propose an approach based on  unconditional image generation models trained on ex-vivo kidney stones; for our experiments, we made use of a Single Image Denoising Diffusion Model (SinDDM) \cite{kulikov2023sinddm}, which is a hierarchical generative model that uses the multi-scale approach of SinGAN \cite{shaham2019singan} combined with the power of denoising diffusion models (DDMs) %\cite{sohldickstein2015deep} 
to generate synthetic images using only images from the training set. %These models make a connection between diffusion probabilistic models and denoising score matching with Langevin dynamics to generate high quality synthetic images \cite{Ho2020Denoising}. The diffusion model consists of two main processes. In the first one,  noise is progressively added to the original image to transform the data distribution to a Gaussian distribution. The generated images are obtained in the reversing process, where denoising is used to restore the initial distribution \cite{Jascha2015Deep}.

The results obtained from this experimental protocol are promising, as they allowed the generation of highly realistic images that were validated experimentally and by experts in the field of MCA which is the standard procedure to determine the kidney stone type. %\cite{daudon2012stone}. 

\section{Related Work}

\subsection{Automatic kidney stone recognition}

Several deep learning (DL) methods have shown promising results in automated kidney stone classification \cite{lopez2024vivo}. However, the effectiveness of DL models is highly dependent on the availability of large datasets for training \cite{lopez2023boosting}. Acquiring such a large dataset poses a challenge in ureteroscopy due to practical limitations \cite{el2022evaluation}.
To address this challenge, techniques such as transfer learning (TL) and fine-tuning from pre-trained models such as ImageNet have been proposed as effective weight initialization strategies \cite{lopez2023boosting}. These methods offer the advantage of avoiding the need to train models from scratch. However, for automated ESR, these initialization techniques may not be applicable, given the substantial disparity between ImageNet distributions and endoscopic (ureteroscopic) images. Consequently, tailored TL methods that initialize weights closer to the target domain are imperative. To address this limitation, Two-Step Transfer Learning has been proposed for kidney stone identification \cite{lopez2023boosting}. Two-step TL allows learning a target distribution (e.g., endoscopic images) using an intermediate distribution (e.g., CCD-camera images).  This technique has shown that using an intermediate domain (similar to the target distribution) substantially improves (up to 10\%) the model over traditional training weights learned from ImageNet \cite{lopez2023boosting}. % The first step of TL consists of training a model using the weights learned from ImageNet (a far distribution of kidney stone images) and fine-tuning with CCD-camera images (near endoscopic images). In a second TL step, the model learns the distribution of the endoscopic image dataset (target domain) and complements it with the weights learned from the distribution of CCD-camera images (intermediate domain) obtained in the first step.
In addition, most existing models for automated ESR have been trained separately on surface or section images \cite{estrade2022towards}. However, visual assessment in MCA (by biologists) and ESR (by urologists) takes advantage of both surface and section fragments, exploiting information from both views\cite{corrales2021classification}.

\subsection{Generative models for data augmentation}
Traditional image data augmentation techniques based on basic image manipulations have been widely used and proven effective in many scenarios but they have limitations in capturing the full diversity and complexity of real-world data \cite{shorten2019survey}.
Advanced methods such as generative models can produce novel and diverse samples, effectively expanding the dataset's diversity. Generative models can be trained on a dataset consisting of examples of the data to be generated and the model learns to capture the patterns and statistical properties present in the training data. This learned knowledge is then used to generate synthetic data with similar characteristics to the original dataset %\cite{foster2019generative}. 
Generative models offer a promising solution to the challenge of scarce data by creating new data samples that closely resemble the original dataset. In scenarios where labeled data is limited or expensive to acquire, generative models can be used to augment the training set, thereby improving model performance and generalization. Different variants of Generative Adversarial Networks (GANs) have been widely used in the medical field for data augmentation %\cite{mcnulty2024synthetic}
\cite{Tiago_2022}, but often have problems with mode collapse and artifact generation \cite{xiaohui2023limited}. The possibility of using Denoising Diffusion Models (DDMs) for medical data augmentation has also been explored, which are more stable models than GANs and have the ability to generate high-quality and diverse synthetic images \cite{xiaohui2023limited}. %\cite{pishva2023repolyp}. 
However, GANs and DDMs require large training datasets, making them inappropriate in cases where a limited training dataset is available. Several studies have proposed the use of different generative models such as SinGAN \cite{shaham2019singan}, %InGAN \cite{shocher2019ingan}, 
MinimalGAN \cite{Zhang2023minimalgan}, among others, to explore their use in generating synthetic images from a small training dataset.

\section{Proposed Approach for Image Generation}
\label{sec:generation}

In order to evaluate the plausibility of synthetically generated images for improving the automated ESR of kidney stones, we follow the pipeline shown in Fig.  \ref{fig:method}. Our hypothesis is that by increasing and balancing the number of images in ex-vivo kidney stone datasets (CCD-camera images) and using synthetic images as intermediate distribution for two-step transfer learning, an improvement over a baseline trained solely on source images can be attained. In this section, the process of synthetic image generation based on CCD-camera images (Fig. \ref{fig:method_generation}) is described in detail. In addition, To determine the performance of the generated images, the evaluation protocol of the synthetic images (Fig. \ref{fig:method_classification}) is proposed in Section \ref{sec:evaluation}.

 The generation of synthetic images (see, Fig. \ref{fig:method_generation}) consists of four fundamental steps: \ref{sec:padding} homogenizing the dimensions of the input set through padding, \ref{sec:sinddm} generating low-resolution synthetic images, \ref{sec:super} resolution enhancement for the synthetic dataset, and \ref{sec:deepchecks} similarity assessment of the distributions of the synthetic images generated with SinDDM and the input set images with DeepChecks.

\subsection{Homogenize image dimensions with padding}
\label{sec:padding}

Often in DL models, images from a training set are required to have homogeneous dimensions. However, in kidney stone image acquisition it is difficult to establish a standard size, as it is highly dependent on acquisition devices in hospitals (such as digital cameras with different capabilities) to perform MCA. Assuming that visual inspection based on MCA is performed on images with multiple resolutions, high quality, controlled acquisition conditions, and a dark background to highlight the colors of kidney stones, padding is applied to all images in the input dataset (Fig. \ref{fig:method_generation}) to homogenize their dimensions. %The dimensions of the input set with padding are 2848$\times$4288 pixels.}

\subsection{Synthetic Image Generation with SinDDM} 
\label{sec:sinddm}

As discussed in Section \ref{sec:intro}, due to the nature of kidney stones, there is a strong class imbalance in the image sets. Therefore, in order to balance and increase the number of samples per class (especially in minority classes), the Single Image Denoising Diffusion Model (SinDDM) \cite{kulikov2023sinddm} is implemented. SinDDM is a generative model that learns the internal statistics of the images on the training set to gradually turn the output image into White Gaussian noise (in a similar way to Denoising Diffusion Models (DDMs)) %\cite{sohldickstein2015deep})
, but hierarchically combining blur and noise. In addition, SinDDM combines the multi-scale approach of SinGAN \cite{shaham2019singan} with the power of DDMs to generate high-quality and diverse synthetic images from a single training image. 

\subsection{Homogening Image Size via Super-Resolution}
\label{sec:super}
The images generated by the SinDDM model have a lower resolution than the original images. For this reason, we used the SwinV2 Transformer for Compressed Image Super-Resolution and Restoration (Swin2SR) model \cite{conde2022swin2sr} to upscale our synthetic images by 4. This is a modified version of SwinIR \cite{liang2021swinir}, which improves the Swin Transformer %\cite{liu2021swin}
abilities in Super-Resolution tasks, particularly in handling Compressed Input SR scenarios. The Swin2SR model uses a traditional upscaling branch utilizing bicubic interpolation, capable of recovering basic structural information. Then, we combine the output of the model with the basic upscaled image to improve its quality.

\begin{figure*} [] 
    \centering
    \subfloat[Dataset A: CCD-camera images (ex-vivo) ]{\label{fig:dataseta}
    \includegraphics[width=0.4\textwidth]{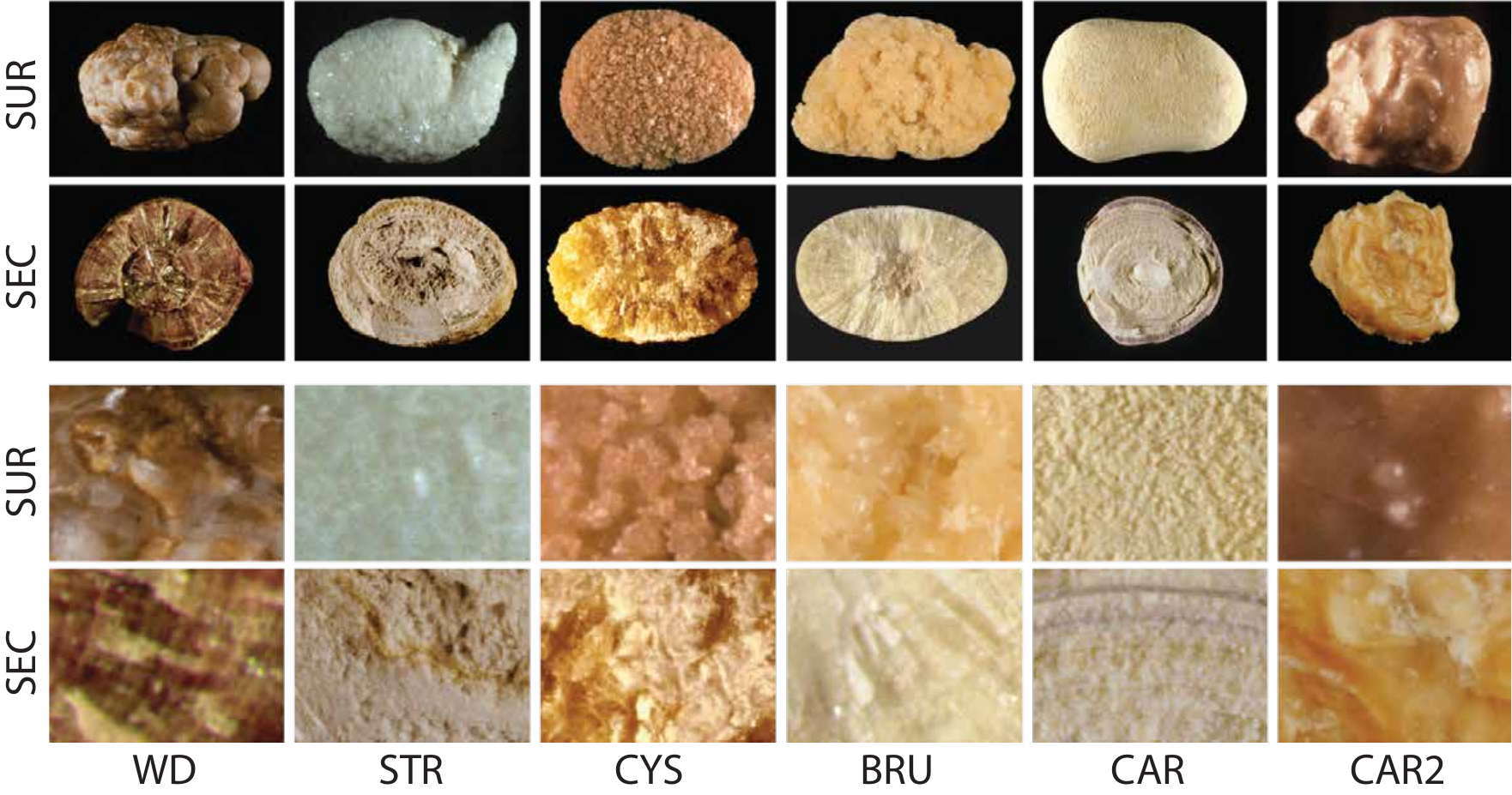}}
    \hspace{5mm}
    \subfloat[Dataset B: Endoscopic images (ex-vivo)]{
\label{fig:datasetb}\includegraphics[width=0.4\textwidth]{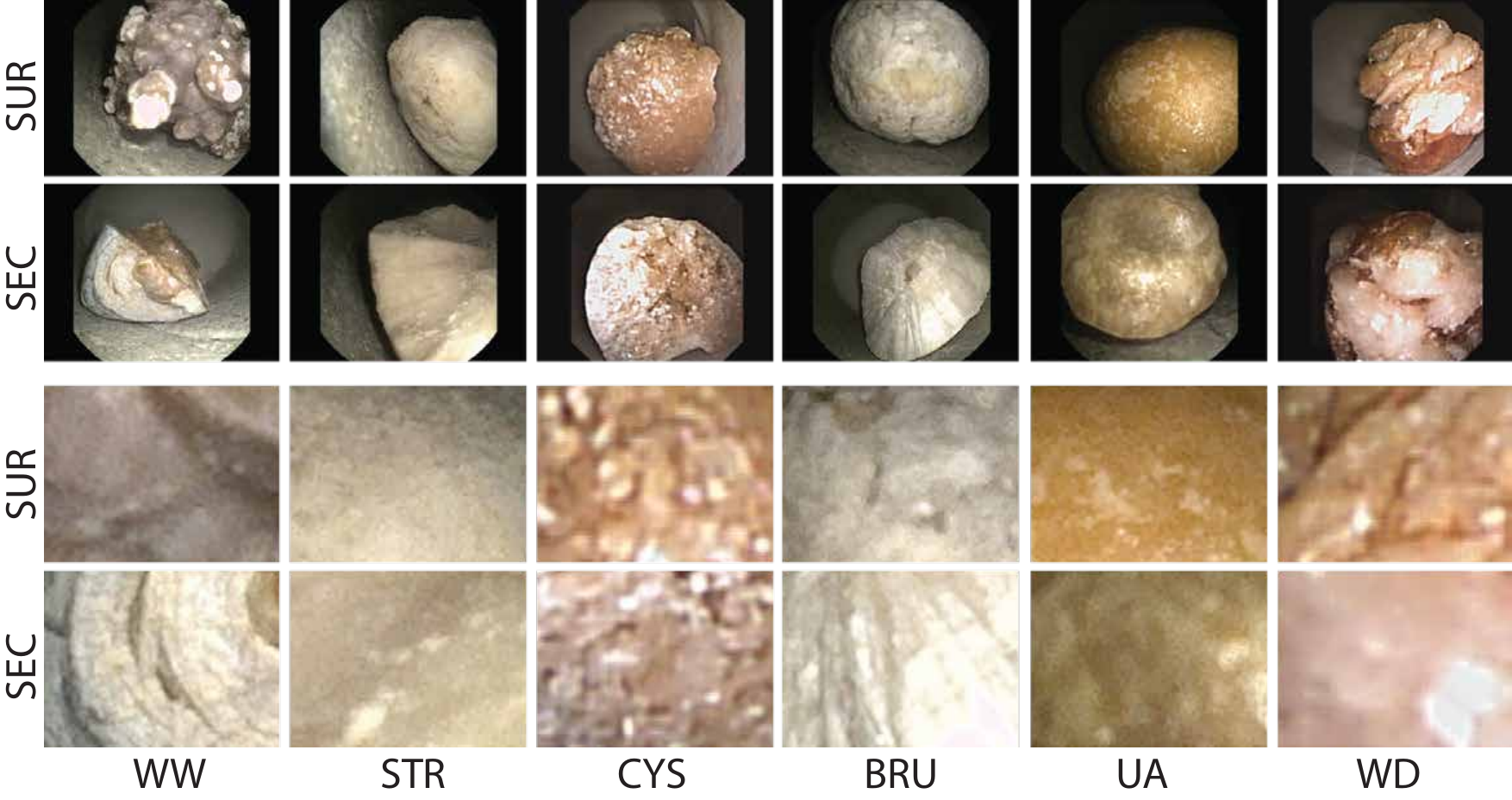}}
\vspace{-0.2cm}
    \caption{Examples of ex-vivo kidney stone images acquired with (a) a CCD camera and  (b) an endoscope. SEC and SUR stand for section and surface views. The first two rows show whole images and in the bottom rows, the sampled image patches} 
    \label{fig:dataset}
    \end{figure*}

\subsection{Quantitative evaluation of generated images}
\label{sec:deepchecks}

\textbf{Deepchecks.} This is a Python library used to test and validate Machine Learning Models and Data \cite{chorev2022deepchecks} and it is composed by checks, conditions, and suites. A check is used to examine a particular property of the data or model. A condition is a function that can be incorporated into a check to assess whether the result meets a predefined standard or criteria. Finally, a suite refers to a structured compilation of checks that can have conditions. Upon execution, a suite generates a detailed report with high-level results and detailed findings.

We used the Train-Test Validation Suite to compare the distributions across our training dataset and the generated images from the SinDDM Model. This Suite contains the following checks: Image Property Drift, Property Label Correlation Change, Label Drift, New Labels, Heatmap Comparison and Image Dataset Drift. For this evaluation, we modified the pre-defined Train-Test suite to obtain a custom suite, which only uses the Heatmap Comparison and the Image Property Drift.
%The Heatmap Comparison creates an average representation of images within each dataset and displays these average images side by side for comparison. This allow us to easily observe differences in brightness distribution between the datasets.
%The second Check used for this evaluation is the Image Property Drift, where different properties are used to extract meaningful information from the data. By examining the distribution of property values, such as identifying extremely dark images or significant difference in contrast or color between the Train dataset and the Synthetic dataset, we can find potential issues in the generative model.
The image properties used by this custom suite are the following: brightness, RMS contrast, mean relative intensity, mean green relative intensity, and mean blue intensity. \\

% \textcolor{red}{Ruben: ¿Puedes volver estos puntos a un parrafo? (para ocupar menos espacio).} 

% \begin{itemize}
%  %   \item \textbf{Aspect Ratio:} Ratio between height and width of image (\(\frac{height}{width})\)
%  %   \item \textbf{Area:} Area of the image in pixels (\(height \times width)\)
%     \item \textbf{Brightness:} Average intensity of the image pixels. Color channels have different weights according to RGB-to-Grayscale formula.
%     \item \textbf{RMS Contrast:} Contrast of image, calculated by standard deviation of pixels.
%     \item \textbf{Mean Red Relative Intensity:} Mean overall pixels of the red channel, scaled to their relative intensity in comparison to the other channels (\(\frac{r}{r + g + b})\).
%     \item \textbf{Mean Green Relative Intensity:} Mean overall pixels of the green channel, scaled to their relative intensity in comparison to the other channels (\(\frac{g}{r + g + b})\).
%     \item \textbf{Mean Blue Relative Intensity:} Mean overall pixels of the blue channel, scaled to their relative intensity in comparison to the other channels (\(\frac{b}{r + g + b})\).
% \end{itemize}

\textbf{Single Image Fréchet Inception Distance (SIFID).} This is a variation of the Fréchet Inception Distance (FID), %\cite{Heusel2018Trained}, 
which is a metric used to evaluate the images generated by generative models. SIFID measures the deviation between the internal distribution of deep features extracted at the output of the convolutional layer just before the second pooling layer.

\subsection{Used Datasets}
\label{sec:datasets}

Three ex-vivo kidney stone datasets were utilized to perform the evaluation protocol. The images were acquired either with standard CCD cameras or with a ureteroscope (i.e., an endoscope), as described as follows in more detail.

\smallbreak 

% Michel Daudon:
\textbf{Dataset A (CCD-camera images).} The ex-vivo dataset described in \cite{corrales2021classification} consists of 366 CCD camera images acquired ex-vivo. These images, as depicted in Fig \ref{fig:dataseta} are divided into 209 surface images and 157 section images. The dataset comprises six different stone types categorized by subtypes denoted by WW (Whewellite, sub-type Ia), CAR (Carbapatite, IVa), CAR2 (Carbapatite, IVa2), STR (Struvite, IVc), BRU (Brushite, IVd), and CYS (Cystine, Va). The acquisition of stone fragment images was carried out using a digital camera under controlled lighting conditions and against a uniform background. %Due to CCD images are acquired with different cameras in different hospitals, the images have different dimensions. 
%To deal with this limitation a padding process was applied to the smaller images so that the dimensions of all the images in dataset A are 2848$\times$4288 pixels. 
The annotation of the images used in this work was statistically confirmed with a study exploiting MCA of extracted kidney stone fragments using microscopy and FTIR analysis.

\smallbreak 

% Synthetic dataset
\textbf{Dataset B (Synthetic images with sinDDM).} 
The synthetic dataset obtained in Section \ref{sec:sinddm}, consists of 300 synthetic images simulating CCD camera images acquired ex-vivo from dataset A. The images were generated with the sinDDM method \cite{kulikov2023sinddm}. These images are divided into two sets of 150 images for surface and section. Each view contains the six subtypes presented in dataset A. Some examples of the generated images are presented in Fig. \ref{fig:motivation}. The dimensions of the images in dataset B are 1056$\times$800 pixels.

% Jonathan El-Beze 
\textbf{Dataset C (Endoscopic images)}. The endoscopic dataset comprises 409 images, as illustrated in Fig. \ref{fig:datasetb}. Within this dataset, there are 246 surface images and 163 section images. Dataset B features the same classes as dataset A, except for substituting Carbapatite fragments (subtypes IVa1 and IVa2) with the Weddelite (subtype IIa) and Uric Acid (IIIa) classes. The images in dataset C were captured using an endoscope, with the kidney stone fragments placed in an environment designed to simulate in-vivo conditions quite realistically (for additional information, refer to \cite{el2022evaluation}). The dimensions of images in dataset B are 576$\times$768 pixels.

%\subsection{Dataset}
% Dataset: Imágenes para generar (Daudon) Menos imágenes. No se ocuparon tantas 

% 

\section{Evaluation protocol}
\label{sec:evaluation}

In order to evaluate the performance of the synthetic images of kidney stones generated in Section \ref{sec:sinddm} with respect to datasets A (CCD-camera images) and C (endoscopic images), two batteries of experiments were performed. The first set of experiments (Section \ref{sec:baseline}) were carried out to create a comparison baseline for each dataset. The aim of the second set of experiments (Section \ref{sec:two-step}) is to assess the plausibility of using synthetic images for training models for endoscopic stone recognition that are able to generalize well to an unseen distribution. For training and testing of both schemes, datasets A, B and C (described in Section \ref{sec:datasets}) are processed as patches (Section \ref{sec:patch}).

\subsection{Patch selection}
\label{sec:patch}
Automatic kidney stone classification typically doesn't analyze full images due to the limited size of datasets \cite{lopez2023boosting}. Instead, patches of 256$\times$256 pixels are often extracted from original images to augment the training dataset size, as detailed in \cite{lopez2024vivo}. % However, it's important to acknowledge the potential drawbacks, such as (i) loss of context, as cropping small regions of an image may result in the loss of contextual and spatial information, and (ii) the possibility of repetitive features across multiple patches. Nevertheless, employing patches offers advantages such as enabling the training of machine learning models, particularly in scenarios with limited samples, and ensuring an increase in sample count and class balance.
A total of 12,000 patches were generated for each dataset, which is categorized into six classes as follows: Datasets A and B include (WW, STR, CYS, BRU, CAR, CAR2), while dataset C includes (WW, WD, UA, STR, BRU, CYS).
A thousand patches are allocated for each class and view (SUR, SEC). However, it is important to note that patches from the same images were excluded from both the training/validation and test datasets.
For each dataset, 80\% of the patches (9600) are utilized for the training and validation phases, while the remaining 20\% (2400) serve as test data (200 patch-images for each class). Patches from the same image are exclusively assigned to either the training/validation or test data.

\subsection{Baseline}
\label{sec:baseline}
In the first set of experiments, three independent models were trained from scratch on datasets A, B, and C, to determine the performance of each model by testing on its distribution (i.e., training on synthetic images, and testing on synthetic images).  See \cite{lopez2023boosting} for implementation details or proceed to ``First Step TL" in the following section.

\subsection{Two-step Transfer Learning}
\label{sec:two-step}

TL has proven to be an efficient technique to adapt a target domain using an intermediate distribution (similar to the target domain), and perform kidney stone classification with acceptable performance \cite{lopez2023boosting, lopez2024vivo}.  In this contribution, two-step TL (Fig. \ref{fig:method_classification}) is adopted to evaluate the performance of datasets A and B (described in Section \ref{sec:datasets}) as an intermediate distribution to classify kidney stones from endoscopic images (Dataset C, target domain).

\textbf{First TL Step: } The first step of TL consists of training a ResNet50 architecture using a large dataset as ImageNet (a far distribution of kidney stone images). Then, a fine-tuning with CCD-camera or synthetic patches (near endoscopic images) and learned weights from ImageNet are used as input to train Model I. The implementation details for the first TL step are as follows: A batch size of 24 was employed alongside an SGD optimizer having a learning rate set to 0.001 and a momentum of 0.9. Fully connected layers incorporating 768, 256, 128, and 6 neurons were introduced, accompanied by batch normalization, ReLU activation, and a dropout rate of 0.5.

\textbf{Second TL Step: } In a second TL step, Model II learns the distribution of the endoscopic image dataset (target domain) and complements it with the weights learned from the distribution of CCD-camera or synthetic images (intermediate domain) obtained in the first step.  The initial weights of Model II are those after Model I is fitted with dataset A or B, and Model II is finally refined with dataset C.
This method aims to enhance the overall generalization capability of Model II and simplify the process of extracting robust features. 
Additionally, 30 epochs were conducted using an SGD optimizer with an increased learning rate of 0.01. Fully connected layers were omitted, as the intention was to utilize the model without any additional alterations to its architecture.

\section{Results and Discussion}

\subsection{Generation}
\label{sec:results_generation}

% Input dataset and padding: 
\textbf{Input dataset and padding:} In order to generate synthetic images from a distribution of CCD-camera images, %\cite{daudon2012stone}),  
eight images per class (see, Section \ref{sec:datasets} ``Dataset A") for each view (SUR and SEC) were used. The images in the training were randomly selected. 
%Often in this type of dataset, multiple images of the same stone are taken at different distances or positions. However, we chose the samples carefully to avoid images being using near duplicates.}
%As the CCD-camera images were acquired in different hospitals with different devices (cameras with different resolution), the dimensions of the images are different. 
A uniform size in the image dimensions is usually required to train the different DL models. Therefore, padding was applied to standardize the dimensions (see Section \ref{sec:padding}). %This is even more convenient, considering that the CCD-camera images used have a dark background, and the padding is done at the edges of the images.
By applying padding to the input images, a training set with 2448$\times$4288 pixel images was obtained. %The criteria for the selected dimensions correspond to the maximum pixel value in rows and columns of the dataset.} 

% Synthetic image generation
\textbf{Synthetic image generation: } For the generation of synthetic images based on CCD-camera images, a total of 12 models were trained (see Fig. \ref{fig:motivation}).
Six models (one for each subtype) for each view (SUR, and SEC) were trained using SinDDM with the same implementation details described in \cite{kulikov2023sinddm}.
Eight CCD-camera images with padding were used as training sets for each model. A total of 25 images were generated by the SinDDM model, which presents similar shape, color, and texture characteristics to the input set. Despite generating synthetic images with high detail preservation, the output dimensions are reduced (264$\times$200 pixels), which is not useful for the classification tasks that will be addressed in Section \ref{sec:evaluation}. 

% Super -> x4
% Synthetic image generation
%\textcolor{blue}{
\textbf{SuperResolution and output dataset: } In order to have synthetic images that can be used in patch-based classification tasks (Section \ref{sec:evaluation}), the synthetic images are processed by a SuperResolution model (Swin2SR \cite{conde2022swin2sr}). The aim of using this model is to increase the image size by a factor of 4 and obtain a set of images with a resolution of 1056$\times$800 pixels.  Although a $\times$8 scale would be more similar to CCD-camera images, we opted for a $\times$4 scale which is more similar to the size of the target domain (images acquired with endoscopes). 

\begin{figure}[t!]
  \begin{center}
    \includegraphics[width=0.45\textwidth]{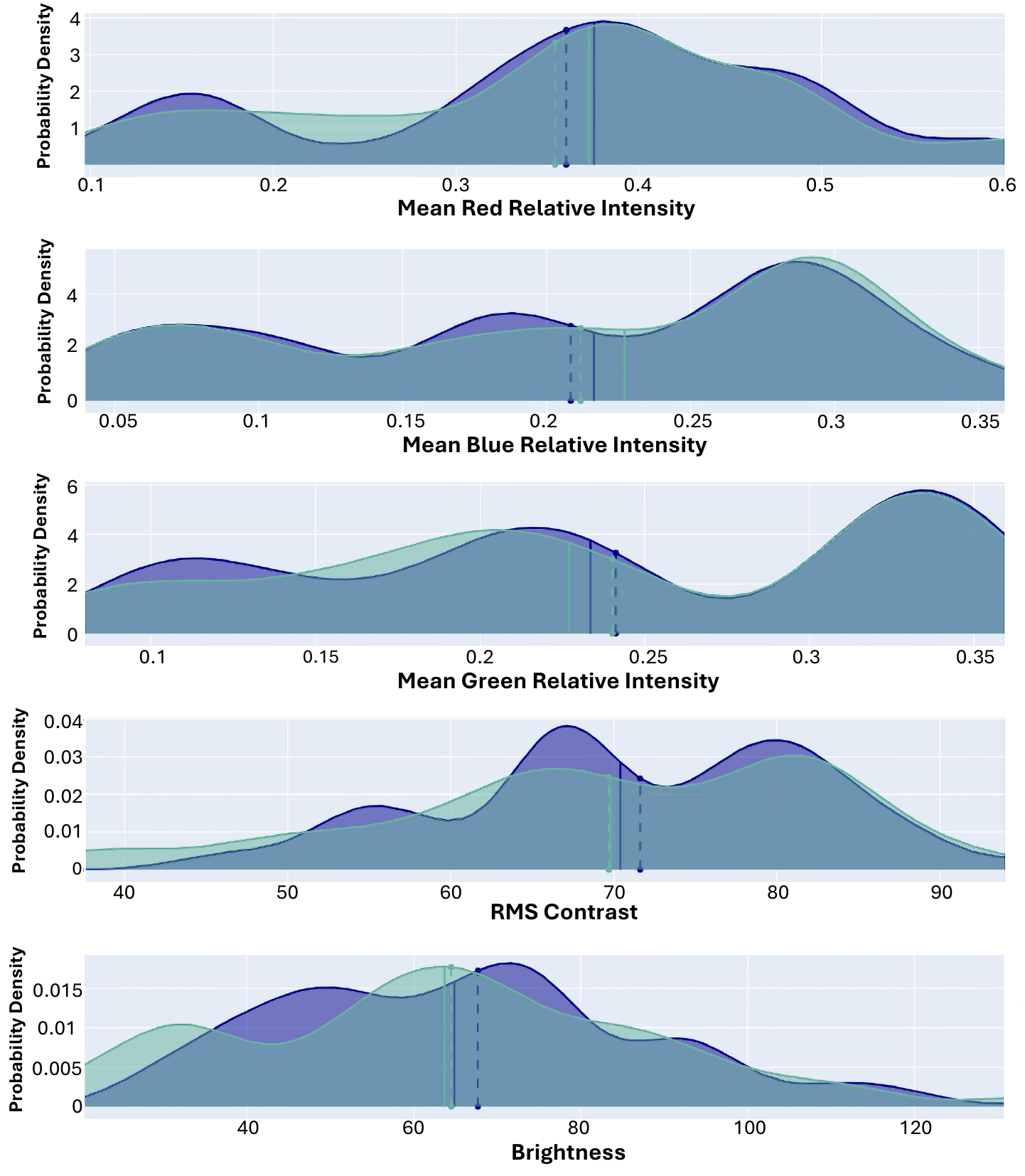}
  \end{center}
  \vspace{-0.3cm}
  \caption{Distribution plot for each image property showing the difference between the train and the synthetic image (SUR view).}
  \label{fig:deepcheck_graphs}
\end{figure}

\textbf{DeepChecks evaluation: } The evaluation of the generated dataset (synthetic images) is performed with DeepChecks against the dataset of CCD-camera images with padding. DeepChecks provided a drift score for 5 characteristics described in Section \ref{sec:deepchecks},  which measures the difference between two distributions, obtained for each image property. A drift score greater than 0.2 means that there is a large difference between the distributions, so the ideal is to obtain a value lower than this limit.
For the SUR view, values of 0.142 brightness, 0.124 RMS Contrast, 0.060 Mean Red Relative Intensity, 0.112 Mean Green Relative Intensity, and 0.094 Mean Blue Relative Intensity were obtained. For the SEC view, values of 0.137 brightness, 0.145 RMS Contrast, 0.077 Mean Red Relative Intensity, 0.055 Mean Green Relative Intensity, and 0.092 Mean Blue Relative Intensity were obtained.   For this evaluation, all the properties obtained a drift score of less than 0.2, which means that the distributions of each of these properties are similar. Figure \ref{fig:deepcheck_graphs} presents the distribution graphs for each evaluated image property, showing us the difference between the distribution of the real images and the distribution of the generated images in the SUR (surface) view.

% \usepackage{booktabs}
% \usepackage{multirow}
\begin{table*}[]
\centering
\caption{Performance comparison (mean ± standard deviation measured as accuracy classification of six different classes) for different Two-Step TL configurations. The baseline is defined to establish a model reference. Two-Step TL* configurations (with a star) denote that the intermediate and target distributions are similar. While a Two-Step TL configuration (without a star) is used to denote an intermediate distribution (such as CCD-camera or synthetic) towards an endoscopic dataset. The best results for each view are denoted in bold. Accuracy I and II, correspond to the test on the 1st and 2nd  step TL, respectively. }
\label{tab:results_classification}
\begin{tabular}{@{}cccccccc@{}}
\toprule
View                 & Initialization & Dataset I              & Dataset II & Model    & Accuracy I & Accuracy II & Configuration \\ \midrule 
\multirow{8}{*}{SUR} & ImageNet       & Synthetic              & No-TL      & ResNet50 & 82.82$\pm$08.72      & No-TL       & Baseline      \\
                     & ImageNet       & CCD-camera             & No-TL      & ResNet50 & 81.72$\pm$09.91      & No-TL       & Baseline      \\
                     & ImageNet       & Endoscopic             & No-TL      & ResNet50 & 68.98$\pm$12.61      & No-TL       & Baseline      \\ \cmidrule(l){2-8} 
                     & ImageNet       & Synthetic              & CCD-camera & ResNet50 & 82.82$\pm$08.72      & 83.81$\pm$04.53       & Two-Step TL*   \\
                     & ImageNet       & CCD-camera             & Synthetic  & ResNet50 & 81.72$\pm$09.91      & 82.62$\pm$03.87       & Two-Step TL*   \\
                     & \textbf{ImageNet}       & \textbf{Synthetic}              & \textbf{Endoscopic} & \textbf{ResNet50} & \textbf{82.82$\pm$04.16}      & \textbf{85.72$\pm$05.75}       & \textbf{Two-Step TL}   \\
                     & \textbf{ImageNet}       & \textbf{CCD-camera}             & \textbf{Endoscopic} & \textbf{ResNet50} & \textbf{81.72$\pm$05.51}      & \textbf{86.19$\pm$06.12}       & \textbf{Two-Step TL}   \\
                     & \textbf{ImageNet}       & \textbf{Synthetic + CCD-camera} & \textbf{Endoscopic} & \textbf{ResNet50} & \textbf{82.34$\pm$04.70}      & \textbf{87.82$\pm$03.65}       & \textbf{Two-Step TL}   \\ \midrule
\multirow{8}{*}{SEC} & ImageNet       & Synthetic              & No-TL      & ResNet50 & 71.53$\pm$10.21      & No-TL       & Baseline      \\
                     & ImageNet       & CCD-camera             & No-TL      & ResNet50 & 71.95$\pm$11.73      & No-TL       & Baseline      \\
                     & ImageNet       & Endoscopic             & No-TL      & ResNet50 & 57.29$\pm$17.87      & No-TL       & Baseline      \\ \cmidrule(l){2-8} 
                     & ImageNet       & Synthetic              & CCD-camera & ResNet50 & 71.53$\pm$10.21      & 72.69$\pm$09.22       & Two-Step TL*   \\
                     & ImageNet       & CCD-camera             & Synthetic  & ResNet50 & 71.95$\pm$11.73      & 72.82$\pm$08.89       & Two-Step TL*   \\
                     & \textbf{ImageNet}       & \textbf{Synthetic}              & \textbf{Endoscopic} & \textbf{ResNet50} & \textbf{73.53$\pm$07.32}      & \textbf{79.95$\pm$05.58}       & \textbf{Two-Step TL}   \\
                     & \textbf{ImageNet}       & \textbf{CCD-camera}             & \textbf{Endoscopic} & \textbf{ResNet50} & \textbf{74.95$\pm$08.58}      & \textbf{78.27$\pm$04.27}       & \textbf{Two-Step TL}   \\
                     & \textbf{ImageNet}       & \textbf{Synthetic + CCD-camera} & \textbf{Endoscopic} & \textbf{ResNet50} & \textbf{75.88$\pm$07.33}      & \textbf{80.76$\pm$04.91}       & \textbf{Two-Step TL}   \\ \bottomrule
\end{tabular}
\end{table*}

\begin{figure}[b!]
  \begin{center}
    \includegraphics[width=0.48\textwidth]{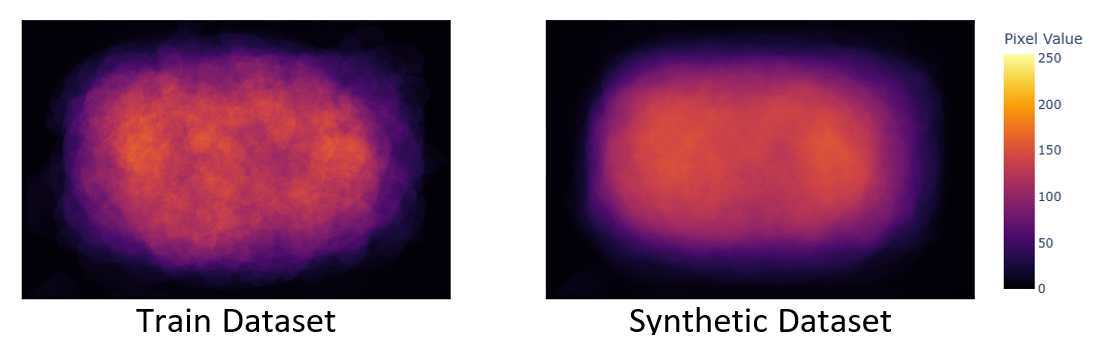}
  \end{center}
  \vspace{-0.3cm}
  \caption{Heatmap Comparison between the Train Dataset and Synthetic Dataset.}
  \label{fig:heatmap}
\end{figure}

Figure \ref{fig:heatmap} shows the Heatmap Comparison, which is an average representation of images within each dataset displayed side by side for comparison. This allows us to easily observe differences in brightness distribution between the datasets and the region in which the dataset images are being generated. As can be seen, most of the images are centered and have a region of interest of up to 50\% of the image. 

\begin{figure}[!b]
  \begin{center}
    \includegraphics[width=0.4\textwidth]{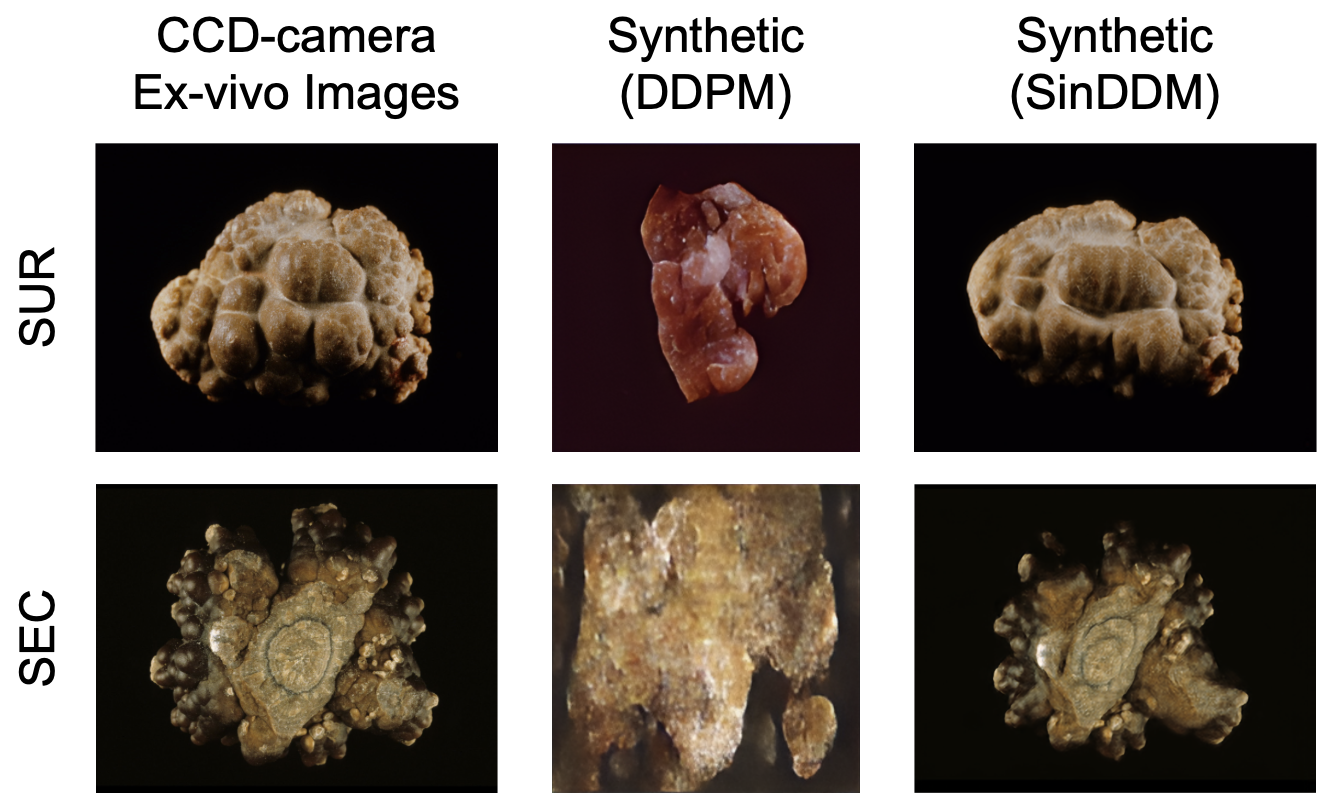}
  \end{center}
  \vspace{-0.3cm}
  \caption{Comparison of the synthetic images generated with DDPM \cite{ho2020denoising} and SinDDM (this contribution) models for the CCD-camera image.}
  \label{fig:comparison}
\end{figure}

Finally, an evaluation was performed with SIFID, a metric to evaluate the similarity between the training set images (CCD images) and the generated set (synthetic images). For the SUR view classes, a SIFID of 3.14$\pm$0.93 was obtained, while for the SEC view it was as high as 3.79$\pm$1.71.
A lower SIFID indicates that the synthetic images are more similar to the training set images. Hence, the SUR images are more similar to the CCD-camera images. In order to qualitatively compare the images generated in this contribution with the state of the art, DDPM model \cite{ho2020denoising} was implemented. Fig. \ref{fig:comparison} shows the comparison of the synthetic images generated with DDPM and SinDDM models concerning the CCD-camera image. As can be seen, the DDPM model lacks colors and textures characteristic of the input images, which presents a high FID (around 90). On the other hand, the SinDDM model presents a high similarity with the training set (a lower SIFID, around 3).

% \begin{figure*} [] % Comprimir al ancho
%     \centering
%     \subfloat[Surface View ]{\label{fig:distribution_SUR}
%     \includegraphics[width=0.48\textwidth]{images/SUR_DC.pdf}}
%     \hspace{3mm}
%     \subfloat[Section View]{
% \label{fig:distribution_SEC}\includegraphics[width=0.48\textwidth]{images/SEC_DC4.pdf}}

%     \caption{Distribution plot for each image property per view showing the difference between the train and the synthetic image. \textcolor{red}{Vamos a eliminar la columna de section. }} 
%     \label{fig:deepcheck_graphs}
%     \end{figure*}

\subsection{Classification}
\textbf{Baseline:} Three models (CCD-camera, synthetic, and endoscopic) based on ResNet50 and ImageNet weights were trained to determine the baseline. As can be seen in Table~\ref{tab:results_classification} (denoted by ``Baseline" in the column ``Configuration"), the performance of the model trained with synthetic images is very similar to the model trained with CCD-camera images. However, the performance of the endoscopic model falls below that of models based on CCD-camera or synthetic images. For SUR, the performance of the models is 82.82$\pm$08.72\% and 81.72$\pm$09.91\% for synthetic and CCD-camera images, respectively. While for the endoscopic dataset, a performance of 68.98$\pm$12.61\% is obtained. On the other hand, for SEC, performance declines as follows: 71.53$\pm$10.21\% and 71.95$\pm$11.73\% for synthetic and CCD-camera, while endoscope barely achieves 57.29$\pm$17.87\%.
However, although the performance of synthetic images is high and similar to CCD-camera images, the real interest lies in enhancing the performance of models trained with endoscopic images (dataset C). For the second set of experiments, two-step transfer learning was performed, to learn in the first step a dataset (CCD-camera or synthetic), and transfer the knowledge to a target distribution such as an endoscopic dataset (Dataset C) as described in Section \ref{sec:two-step}.

\textbf{Two-step Transfer Learning: } To ascertain the performance between the synthetic and CCD-camera datasets, and vice versa, the two-step TL* (with star) configuration was used (see Table \ref{tab:results_classification}).  For both views, it can be observed that performing TL between similar distributions does not significantly improve the training (similar to the performance obtained in baseline). This could be because there is no more relevant color or texture information to be learned from the distribution. 
Finally, to evaluate the performance of the synthetic images concerning the CCD-camera as an intermediate distribution, Two-Step TL (no star) is performed towards the endoscopic dataset. In addition, a further test is performed by combining the images from the CCD-camera and synthetic distribution, to determine if there is any positive effect when combining both sets. The results for the SUR view suggest an improvement from 82.82$\pm$04.16\% to 85.72$\pm$05.75\% with the synthetic to endoscopic configuration. While the CCD-camera to Endoscopic configuration improves from 81.72$\pm$05.51\% to 86.19$\pm$06.12\%. For the third configuration (Synthetic+CCD-camera to Endoscopic), there is a significant increase from 82.34$\pm$04.70\% to 87.82$\pm$03.65\%. On the other hand, for the SEC view, they follow a similar trend from 73.53$\pm$07.32\% to 79.95$\pm$05.58\% in the synthetic to Endoscopic configuration. Also, there is an increase from 74.95$\pm$08.58\% to 78.27$\pm$04.27\% from CCD-camera to Endoscopic, respectively. Finally, for Synthetic to CCD-camera to Endoscopic distribution, there is an increase from 75.88$\pm$07.33\% to 80.76$\pm$04.91\%.

\section{Conclusions and Future Work}
The identification of kidney stones using machine learning techniques is an open problem. One of the main limitations is the lack of data to train models, especially when there is a class imbalance, which is common in the context of kidney stones. In this work, we proposed the generation of synthetic images to balance and increase the number of samples in datasets. Furthermore, the generation of synthetic CCD-camera images was evaluated using DeepChecks, which shows that the characteristics such as color and texture are similar between the two sets. Subsequently, a Two-Step Transfer Learning method was implemented to evaluate the automatic classification performance on endoscopic images using the synthetic images as an intermediate domain. It was demonstrated that synthetic images are useful for achieving performance similar to CCD-camera images. Combining both sets represents a good starting point for evaluating endoscopic images.

In future work, it is desired to generate and classify whole CCD-camera or endoscopic images. As well as, to control the percentage of the kidney stone with respect to the image size (useful for patch-based classification). In addition, it is expected to generate synthetic endoscopic images to train models with real artifacts in clinical practice (such as illumination changes and blur), and also compare with models such as Generative Adversary Networks.  Perhaps the greatest challenge will be to insert kidney stones that appear on CCD-camera images in an endoscopic context, where the surrounding tissue is observed and not a dark background.

\section*{Acknowledgements}
The authors wish to acknowledge the Mexican Council for Humanities, Science, and Technology (CONAHCYT) for their support in terms of postgraduate scholarships in this project, and the Data Science Hub at Tecnologico de Monterrey for their support on this project. 
This work has been supported by Azure Sponsorship credits granted by Microsoft's AI for Good Research Lab through the AI for Health program. The project was also supported by the French-Mexican ANUIES CONAHCYT Ecos Nord grant (MX 322537/FR M022M01).

\bibliography{egbib}{}
\bibliographystyle{IEEEtran}
\vspace{12pt}

\end{document}